\documentclass[conference]{IEEEtran}
\IEEEoverridecommandlockouts
\usepackage{cite}
\usepackage{amsmath,amssymb,amsfonts}
\usepackage{algorithmic}
\usepackage{fancyhdr}
\usepackage{graphicx}
\usepackage{textcomp}
\usepackage{xcolor}
\usepackage{tabularx}
\usepackage{float}
\usepackage{caption} 
\usepackage{array} 

\def\BibTeX{{\rm B\kern-.05em{\sc i\kern-.025em b}\kern-.08em
    T\kern-.1667em\lower.7ex\hbox{E}\kern-.125emX}}

\begin{document}

\title{SynthEnsemble: A Fusion of CNN, Vision Transformer, and Hybrid Models for Multi-Label Chest X-Ray Classification\\
}

\author{
\IEEEauthorblockN{S.M. Nabil Ashraf, Md. Adyelullahil Mamun, Hasnat Md. Abdullah, Md. Golam Rabiul Alam}
\IEEEauthorblockA{\textit{Department of Computer Science and Engineering} \\
\textit{BRAC University, 66 Mohakhali, Dhaka - 1212, Bangladesh} \\
\{s.m.nabil.ashraf, md.adyelullahil.mamun, hasnat.md.abdullah\}@g.bracu.ac.bd, rabiul.alam@bracu.ac.bd}
}
\maketitle
\thispagestyle{fancy}

\begin{abstract}
 Chest X-rays are widely used to diagnose thoracic diseases, but the lack of detailed information about these abnormalities makes it challenging to develop accurate automated diagnosis systems, which is crucial for early detection and effective treatment. To address this challenge, we employed deep learning techniques to identify patterns in chest X-rays that correspond to different diseases. We conducted experiments on the "ChestX-ray14" dataset using various pre-trained CNNs, transformers,  hybrid(CNN+Transformer) models, and classical models. The best individual model was the CoAtNet, which achieved an area under the receiver operating characteristic curve (AUROC) of 84.2\%. By combining the predictions of all trained models using a weighted average ensemble where the weight of each model was determined using differential evolution, we further improved the AUROC to 85.4\%, outperforming other state-of-the-art methods in this field. Our findings demonstrate the potential of deep learning techniques, particularly ensemble deep learning, for improving the accuracy of automatic diagnosis of thoracic diseases from chest X-rays. Code available at: https://github.com/syednabilashraf/SynthEnsemble
\end{abstract}


\section{Introduction}
The field of medical diagnostics has witnessed a growing interest and recognition because of deep learning as a promising and feasible approach. Specifically, using chest X-ray imaging as a screening and diagnostic modality in Artificial Intelligence tools holds significant importance for various thoracic diseases \cite{rajadanuraks_performance_2021}. However, the lack of properly labeled hospital-scale datasets, as well as  fine-grained features, are hindering the development of computer-aided diagnosis systems\cite{wang_chestx-ray8_2017}. Despite this, the utilization of Convolutional Neural Networks (CNNs), pre-trained transformer models, and their subsequent fine-tuning for downstream tasks has demonstrated efficacy in situations where there is a scarcity of training data and quality features \cite{seyyed-kalantari_chexclusion_2020} \cite{woo_convnext_2023}. Additionally, it is imperative to mitigate unexpected biases, as they are deemed undesirable within a medical scenario. In the context of low-resolution images and limited image data, it was seen that Swin Transformer V2 outperforms alternative vision transformer models\cite{liu_swin_2021}.

This study investigates various deep-learning approaches for the purpose of identifying features in chest radiography (CXR) pictures that are indicative of chest illnesses. In every instance of chest disease, we employ pre-trained convolutional neural network (CNN) models, vision transformer models, and a fusion of CNN and transformers, utilizing chest X-ray pictures as input. Furthermore, we optimize the hyper-parameters of the multi-label system, which encompasses 14 diagnostic labels concurrently.\\
In our study, we present three noteworthy contributions:

\begin{itemize}
    \item 
We outperformed previous efforts in thoracic disease identification by achieving a superior ROC AUC. To achieve this, we conducted a thorough comparison of diverse neural network models, including transformers, Convolutional Neural Networks (CNNs), hybrids (CNN + ViT), and even classical models.
    \item 
We developed an innovative approach that involves the seamless integration of diverse Deep Neural Networks and Classical Machine Learning Models. By incorporating Ensemble learning techniques, we effectively elevated the performance of the model.
    \item 
    We utilized a unique approach to make the training more cost-effective by using a cyclic learning rate.
     Moreover, we implemented a two-step fine-tuning and training process, accompanied by the identification of optimal learning rates for each step. This approach was employed to achieve improved and expedited convergence. Collectively, these contributions shed light on a pathway for advancing within the realm of thoracic disease identification.
\end{itemize}


\section{Literature Review}



In recent years, significant progress has been made in the field of deep learning and the availability of extensive datasets in the task of medical imaging. These advancements have facilitated the development of methods that have demonstrated comparable performances against healthcare experts in various medical imaging tasks\cite{gulshan_development_2016} \cite{esteva_dermatologist-level_2017} \cite{rajpurkar_cardiologist-level_2017} \cite{grewal_radnet_2018} \cite{lakhani_deep_2017}. In particular, detection of diabetic retinopathy\cite{gulshan_development_2016}, classification of skin cancer\cite{esteva_dermatologist-level_2017}, identifying arrhythmia\cite{rajpurkar_cardiologist-level_2017}, recognition of haemorrhage\cite{grewal_radnet_2018}, and pulmonary tuberculosis detection in x-rays\cite{lakhani_deep_2017}

In continuation of expanding the realm of this medical imaging field, Wang et al.\cite{wang_chestx-ray8_2017} introduced the ChestX-ray-14 dataset, which contains significantly larger data compared to prior datasets in the same domain. Additionally, Wang et al. conducted a comparative evaluation of various convolutional neural network architectures that had been pre-trained on ImageNet\cite{russakovsky_imagenet_2015}. After that, researchers have come forward to improve the detection of different chest diseases by proposing and leveraging different methodologies, like Yao et al.\cite{yao_learning_2018}, who tested the statistical dependencies among labels.

\subsection{Convolutional Neural Networks in medical image domain}

In the field of medical image learning, it is not unexpected to leverage the Convolutional Neural Network (CNN)\cite{lecun_gradient-based_1998} as the foundation for the most successful models in the field of medical image learning. Among researchers, the CNN structure has been the prevailing choice for image recognition tasks. As a result, many have proposed efficient methodologies on top of the existing ones. Gao et al. \cite{huang_densely_2016} illustrated a solution to the vanishing-gradient problem in Convolutional Neural Networks (CNNs) by interconnecting each layer of the CNN with every subsequent layer. However, the complex architectures of CNNs give rise to concerns regarding interpretability and computational efficiency. Nevertheless, Pranav et al. successfully employed 121-layer CNNs to address medical imaging challenges in the realm of Cardiovascular diseases, demonstrating consistent performances \cite{rajpurkar_chexnet_2017}.

\subsection{Transformer in the medical computer vision}
In recent times, in the context of the classification and detection task, self-supervised learning techniques such as masked autoencoders\cite{he_masked_2021} are being used to improve the performance of pure CNNs. Post-2020, the adoption of the Transformer model in computer vision has become evident, attributed to its significant capability, as outlined in Vaswani et al.'s work \cite{vaswani_attention_2023}. Nevertheless, for effective utilization of transformers, self-attention, and self-supervision techniques in image processing, various researchers have suggested diverse enhancements. Zihang et al. \cite{dai_coatnet_2021} combined the strengths of transformers and convolutional networks, emphasizing the synergies between these two architectures. They also brought attention to the issue of limited scalability in the self-attention mechanism of transformers when dealing with larger image sizes. Addressing this concern, Li et al. \cite{yuan_volo_2021}, in their paper on Vision Outlooker (VOLO), pointed out its effectiveness in encoding fine-grain features and contextual information into tokens—an achievement not previously attainable through self-attention mechanisms. Additionally, this study led to the development of MaxVIT \cite{tu_maxvit_2022} to better accommodate larger image sizes.

While both CNNs and Transformers contribute significant roles in medical computer vision, they exhibit distinct strengths and weaknesses. CNNs excel in scalability and have demonstrated high performance on large datasets. Conversely, transformers, such as ViTs, introduce innovative approaches with attention mechanisms. However, addressing the scalability limitations of transformers remains a considerable challenge, which this study tries to overcome. We have attempted to harness the capabilities of recent ViT models like Swin Transformer v2 \cite{liu_swin_2021}, initially trained on low-quality images, to effectively handle downstream tasks involving higher-resolution images and mitigate the scalability issues.



\section{Methodology}

\subsection{Dataset Description}

Our research uses the ChestX-ray14 dataset\cite{wang_chestx-ray8_2017}, a robust compilation comprising 112,120 frontal-view X-ray images from 30,805 unique patients from 1992 to 2015. Expanding on the ChestX-ray8 dataset, this comprehensive collection incorporates six additional extrathoracic conditions, including Edema, Emphysema, Fibrosis, Pleural Thickening, and Hernia.

\begin{figure}[H]
    \centering
    \includegraphics[width=0.9\linewidth]{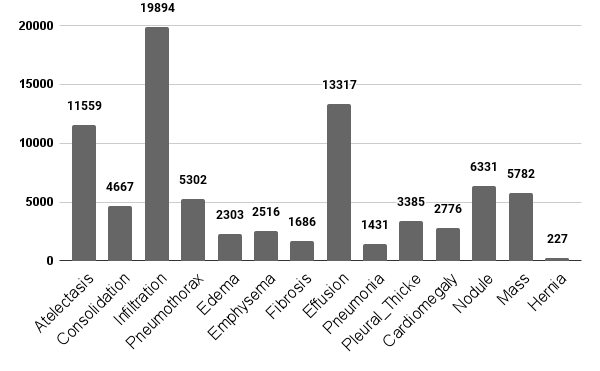}
    \caption{Disease distribution in the dataset.}
    \label{fig:data-distribution}
\end{figure}

Figure \ref{fig:data-distribution} illustrates the inherent class imbalance within our dataset, revealing that certain medical conditions are disproportionately represented. This imbalance could potentially lead to biased model performance. Moreover, a considerable portion of the dataset, amounting to more than 60,000 (out of 112,120 images), was categorized as "No Findings," indicating the absence of any of the 14 detectable diseases.

\subsection{Model Exploration}\label{AA}
In this section, we briefly introduce different types of cutting-edge image classification models that we have selected for our experiment, including CNN, Vision Transformer (ViT), and Hybrid (CNN + ViT) models.

\subsubsection{Hybrid architectures}

\begin{itemize}
    \item CoAtNet: CoAtNets \cite{dai_coatnet_2021} comprise a novel class of hybrid models that seamlessly merge depthwise convolution and self-attention through simple relative attention mechanisms. This approach leads to a coherent fusion of convolution and attention layers, effectively enhancing generalization, capacity, and efficiency. Furthermore, CoAtNets demonstrate improved performance by systematically stacking convolutional and attention layers in a meticulously designed manner.

    \item MaxViT: MaxViT\cite{tu_maxvit_2022} is a pioneering model that leverages multi-axis attention to achieve powerful global-local spatial interactions on diverse input resolutions, all while maintaining linear complexity. By ingeniously incorporating blocked local and dilated global attention mechanisms, MaxViT empowers the integration of attention with convolutions. This ingenious synergy culminates in a hierarchical vision backbone, where the fundamental building block is seamlessly replicated across multiple stages.
\end{itemize}

\subsubsection{Vision Transformer  (ViT)}

\begin{itemize}
    \item Swin V2: Swin Transformer V2\cite{liu_swin_2021} introduces an array of strategies, including residual-post-norm with cosine attention, log-spaced position bias, and SimMIM self-supervised pre-training. These techniques collectively foster training stability, resolution transfer, and reduction in labeled data dependency. The outcome is a  model that achieves remarkable performance across diverse vision tasks, surpassing state-of-the-art benchmarks. 

    \item VOLO: Addressing the limitations of self-attention, Vision Outlooker (VOLO)\cite{yuan_volo_2021} introduces outlook attention, an innovative technique that efficiently captures fine-grained features and contexts at a more granular level. This novel approach, embodied in the VOLO architecture, stands in contrast to conventional self-attention, which predominantly focuses on coarser global dependency modeling. 
\end{itemize}

\subsubsection{Convolutional Neural Network (CNN)}

\begin{itemize}
    \item DenseNet: Dense Convolutional Networks (DenseNets)\cite{huang_densely_2016} present a departure from conventional architectures by establishing dense connections that link each layer to all other layers in a feed-forward manner. This unique connectivity pattern results in an exponential increase in direct connections, mitigating challenges associated with vanishing gradients and facilitating robust feature propagation. DenseNets also engender feature reuse and parameter reduction, showcasing their efficacy in optimizing image classification tasks.

    \item ConvNeXt V2: Building upon the ConvNeXt framework, ConvNeXt V2\cite{woo_convnext_2023} introduces an upgraded model with a fully convolutional masked autoencoder structure and a Global Response Normalization (GRN) layer. This integration of self-supervised learning techniques and architectural refinements contributes to substantial performance enhancements across various recognition benchmarks, underscoring the potency of combined approaches in image classification.
\end{itemize}

\subsection{Data pre-processing and splitting}
This section outlines the fundamental steps taken to pre-process the data, ensuring its suitability for subsequent analysis and model training.
\subsubsection{Image Resizing and Normalization}
Initially, the images were sized 1024x1024 pixels, and we resized them to more manageable dimensions of 224x224 pixels to enhance computational efficiency within resource constraints. We normalized the images using a mean and standard deviation of images from the Imagenet dataset.

\subsubsection{Horizontal Flips and Rotation}
 We incorporated random horizontal flips and rotations to enhance orientation robustness and promote feature learning. These augmentations were applied with a 50\% probability each, and rotations were confined to a maximum of 10 degrees.

\subsubsection{Splitting}
The dataset was divided into distinct groups, with 70\% allocated for training, 20\% for testing, and 10\% for validation. Notably, patient overlaps were meticulously avoided across these divisions, as evident in Table \ref{tab:splitting}. As indicated by Yao et al.\cite{yao_learning_2018}, variations in random splitting negligibly impact performance, thus guaranteeing an equitable basis for comparison.

\begin{table}[h]
    \centering
        \captionsetup{justification=centering} 
    \caption{SPLITTING OF THE DATASET}
    \begin{tabular}{|c|c|c|c|c|}
        \hline
         & Total & Train & Validation & Test \\
        \hline
        Images & 112120 & 78544 & 11220 & 22356 \\
        \hline
        Unique Patients & 30805 & 21563 & 3081 & 6161 \\
        \hline
    \end{tabular}
    \label{tab:splitting}
\end{table}

\subsection{Training and Optimization}

\begin{itemize}
    \item \textbf{Finding initial learning rate}: To identify the initial learning rate (LR), we leverage the Learning Rate Range Test, a technique discussed by Smith\cite{smith_cyclical_2015}. The crux of this approach is centered around the concept of cyclical learning rate (CLR), which alternately increases and decreases during the training process. Our choice of the optimizer is AdamW\cite{loshchilov_decoupled_2017}, which we adopt with CLR during training. 
We ran a small training session for 100 iterations in which the learning rate was increased between two boundaries, $\text{min\_lr}$ and $\text{max\_lr}$, which were 1e-7 and 1e-1, respectively. We then plot the LR vs. Loss curve and pick the midpoint of the steepest descending portion of the curve as the maximum bound for training with CLR.

\begin{figure}
    \centering
    \includegraphics[width=0.8\linewidth]{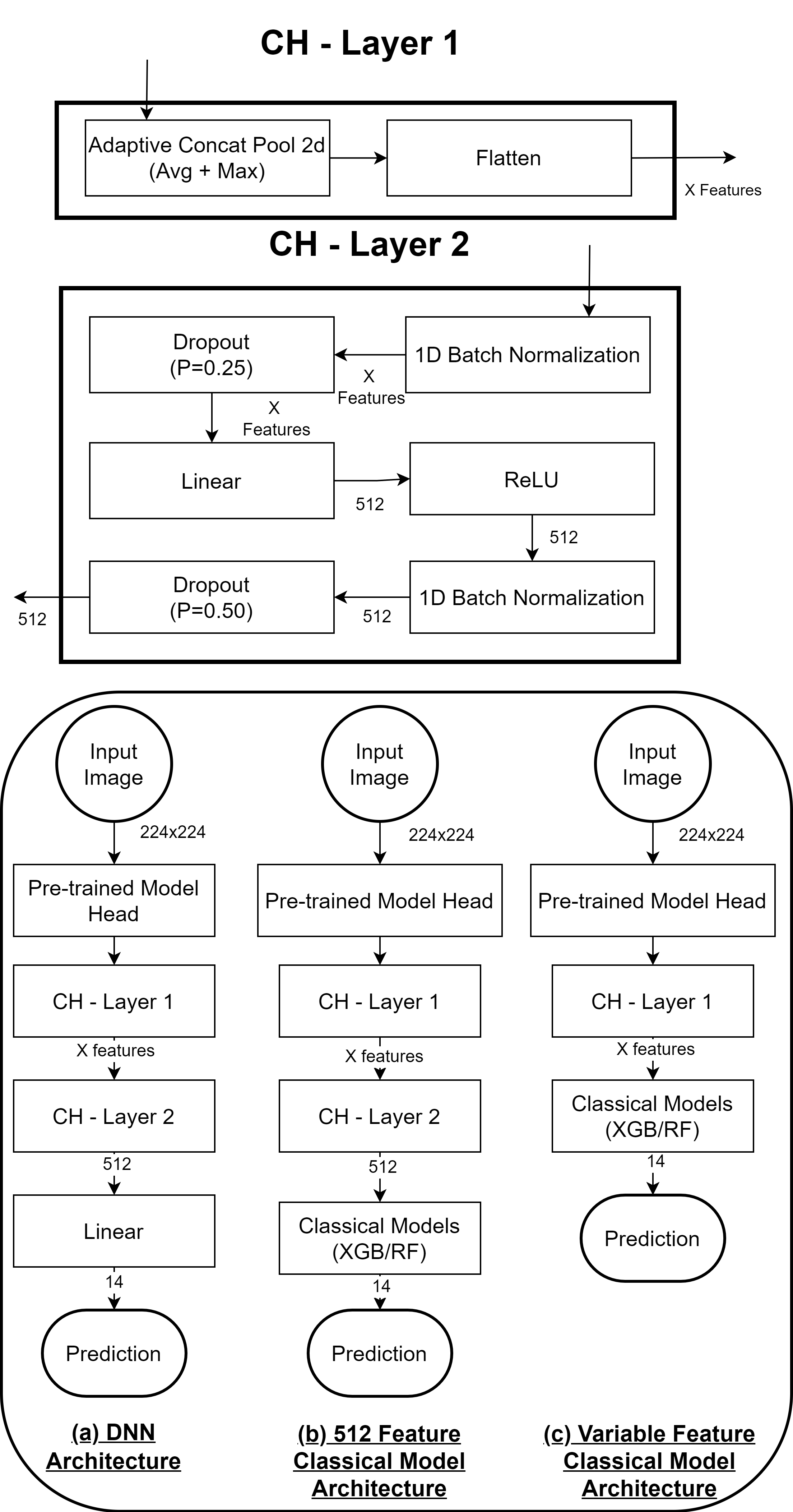}
    \caption{Architecture for DNN and classical model. CH: Custom Head. XGB: XGBoost. RF: Random Forest}
    \label{fig:dnn-classical-architecture}
\end{figure}
\begin{figure}
    \centering
    \includegraphics[width=0.8\linewidth]{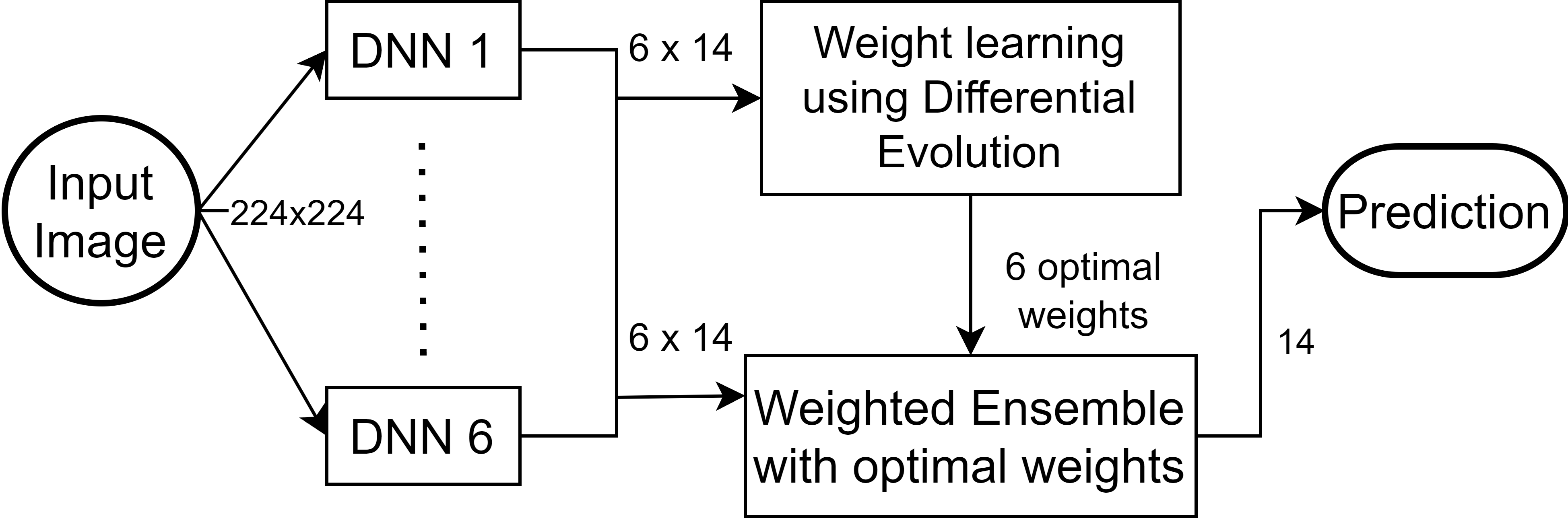}
    \caption{Average weighted ensemble with differential evolution.}
    \label{fig:ensemble}
\end{figure}

    \item \textbf{Training DNN}: Figure \ref{fig:dnn-classical-architecture}a  illustrates the architecture used for training every DNN. We employed pre-trained weights sourced from ImageNet for initializing each neural network. The model's head was trained for three epochs, as it was randomly initialized. Then, the full network was fine-tuned for ten epochs with discriminative LR, where the model’s initial layers were  trained with a lower LR compared to the final layers. Training was halted when the model failed to improve for three consecutive epochs. The best model with the lowest validation loss during training was saved. To optimize the initial learning rate for the fine-tuned model, the weights of the saved model were loaded, and the LR Range Test was re-run. The entire model was then trained for a second time using the optimal initial learning rate for five epochs, saving only the model with the best validation loss. While training in the second phase only slightly improved the validation loss for some models like CoAtNet, all six models were trained using the same approach with two phases.
    
    \item \textbf{Training Classical Models as Meta-Learner}: To enhance outcomes, we explored utilizing feature vectors generated by the top-performing DNN as input for classical models like XGBoost and Random Forest. We pursued two strategies: firstly, training classical models with the output vectors from the second-to-last layer of our DNN models (varied per model) as depicted in Figure \ref{fig:dnn-classical-architecture}c; secondly, training them with the last layer of our custom head, producing 512 features for each DNN as shown in Figure \ref{fig:dnn-classical-architecture}b. Integrating the top-performing DNN with classical models aimed to synergize model strengths and enhance overall performance.
    \item \textbf{Ensembling DNN}: To make our predictions more robust with reduced variance, we used two different ensemble techniques on the validation split before evaluating on test data. (1) Stacking:  we concatenated the six probability vectors outputted by all 6 DNNs to form 84 features (6 DNN models * 14 probabilities each = 84 features) for each image and trained XGBoost as a meta classifier to make the final predictions. (2) Weighted average: we averaged the six probability vectors using different weights for each DNN to produce one probability vector for each image. The optimal weight for each model, which determined its contribution to the weighted final prediction, was found using a stochastic global search algorithm known as differential evolution\cite{differential_evolution}. The weights were bounded 0 and 1 (inclusive) and summed to 1. This was the superior ensemble technique and has been shown in Figure \ref{fig:ensemble}. 

\end{itemize}

\section{Experiments}

\subsection{Experimental Setup}

We chose PyTorch as our implementation platform. Our experiments were run on two Nvidia T4 GPUs with 16GB of memory each. We selected Binary cross-entropy loss as our loss function for multi-label classification. We employed AdamW as our optimizer with a true weight decay of 1e-2 for every DNN and a momentum of 0.9. We set the batch size to 32, which used the full capacity of our GPUs. Our training process incorporates an early stopping mechanism, stopping training if the validation loss does not improve by a margin of 1e-3 within five epochs. Additionally, to safeguard model progress, we implement a checkpoint system, preserving the model's state each time the validation loss experiences a reduction of at least 1e-4.

\subsection{Models Performance}

Among all of our DNN models, CoAtNet performed the best among all models with a mean AUROC of 84.2\%, followed by ConvNeXtV2 with 84.1\% (shown in Table \ref{tab:dnn_model_summary}).

\begin{table}[H]
    \centering
        \caption{AUROC of DNN models}

    \begin{tabular}{|l|l|l|r|}
        \hline
        Model & Type & Model Param (M) & AUROC \\
        \hline
        CoAtNet & Hybrid & 73.9 & \textbf{0.84239} \\
        \hline
        MaxViT & Hybrid & 116 & 0.84013 \\
        \hline
        DenseNet121 & CNN & 8 & 0.82440 \\
        \hline
        ConvNeXtV2 & CNN & 198 & 0.84091 \\
        \hline
        VOLO & Transformer & 58.7 & 0.83205 \\
        \hline
        SwinV2 & Transformer & 49.7 & 0.83573 \\
        \hline
    \end{tabular}
    \label{tab:dnn_model_summary}
\end{table}


Next, we used the feature vectors from these models to train classical models (XGBoost and Random Forest), as detailed in Table \ref{tab:classical_model_summary}. Notably, CoAtNet features surpassed ConvNeXtV2 in both cases (last and second-to-last layers), with clear performance improvement using 512 features compared to the larger alternative.

\begin{table}[H]
    \centering
    \caption{AUROC OF CLASSICAL MODELS}
    \begin{tabular}{|l|l|l|l|}
        \hline
        Feature Extraction Model &Classifier Model &Features & AUROC \\
        \hline
        CoAtNet &XGB & 512 & \textbf{0.83814} \\
        \hline
        CoAtNet & XGB &2048 & 0.82354 \\
        \hline
        ConvNeXtV2 & XGB &512 & 0.82536 \\
        \hline
        ConvNeXtV2 & XGB &3072 & 0.81160 \\
        \hline
        CoAtNet & RF & 512 & 0.81441 \\
        \hline        
        CoAtNet& RF& 2048 & 0.80607 \\
        \hline        
        ConvNeXtV2& RF& 512 & 0.80150 \\
        \hline
        ConvNeXtV2& RF& 3072 & 0.79458 \\
        \hline
    \end{tabular}
    \label{tab:classical_model_summary}
\end{table}

Lastly, we ensembled all 6 DNNs with two distinct techniques. (1) Stacking: the probability vectors outputted by each DNN to train XGBoost as a meta-classifier. (2) Unweighted and weighted average ensemble where the weights were determined using differential evolution. We assessed each technique on validation split  before evaluating on test data, as shown in Table \ref{tab:ensemble_combinations}. The weighted average ensemble demonstrated superior performance overall, outperforming individual DNN models as well as other ensemble methods.

\begin{table}[H]
    \centering
    \caption{AUROC WITH DIFFERENT ENSEMBLE TECHNIQUES}
    \begin{tabular}{|l|c|}
        \hline
        Technique &  AUROC \\
        \hline
         Stacking with Meta Classifier (XGB) & 0.84518 \\
        \hline
         Unweighted Average Ensemble & 0.85327 \\
        \hline
        Weighted Average Ensemble & \textbf{0.85433} \\
        \hline

    \end{tabular}
    \label{tab:ensemble_combinations}
\end{table}

\subsection{Comparison of DNN, ML and Ensemble model}
Our experimental investigation was structured around three distinct categories: Deep Learning Models, Classical Models with deep learning model features, and an ensemble approach that combines all models. Within the Deep Learning Models category, we further categorized models into CNN, Transformer, and Hybrid (CNN + Transformer) architectures, as shown in Table \ref{tab:ensemble_classical_dnn}. Our selection process involved identifying the most promising model from each category within Deep Neural Network (DNN), classical Machine Learning (ML), and Ensemble models.

Among six evaluated DNN models (CoAtNet, MaxViT, DenseNet121, ConvNeXtV2, VOLO, and SwinV2), CoAtNet performed the best. We used CoAtNet and ConvNeXtV2's feature vectors for training classical ML models (XGBoost, Random Forest), but they fell short of deep learning models' performance. Instead, an ensemble model combining all models consistently outperformed others, emphasizing its predictive accuracy and robustness. In summary, CoAtNet excelled in DNNs, classical models couldn't match deep learning models' performance, and the ensemble model dominated predictive accuracy across all labels.

\begin{table}[H]
    \centering
    \caption{AUROC OF BEST DNN, CLASSICAL AND ENSEMBLE MODEL FOR ALL DISEASES}
    \begin{tabular}{|l|c|c|c|}
        \hline
        Disease & CoAtNet & CoAtNet+XGB & Weighted Ensemble \\
        \hline
        Atelectasis & 0.82313 & 0.82364 & \textbf{0.83390} \\
                \hline
        Consolidation & 0.80980 & 0.81151 & \textbf{0.81575} \\
                \hline
        Infiltration & 0.73105 & 0.73199 & \textbf{0.74102} \\
                \hline
        Pneumothorax & 0.89660 & 0.89068 & \textbf{0.90164} \\
                \hline
        Edema & 0.90185 & 0.90214 & \textbf{0.91034 }\\
                \hline
        Emphysema & 0.92067 & 0.91891 & \textbf{0.92946} \\
                \hline
        Fibrosis & 0.81574 & 0.80103 & \textbf{0.83347} \\
                \hline
        Effusion & 0.88203 & 0.88147 & \textbf{0.88977} \\
        \hline
        Pneumonia & 0.76093 & 0.75798 & \textbf{0.77648} \\
        \hline
        Pleural\_Thickening & 0.80053 & 0.80448 & \textbf{0.81270} \\
        \hline
        Cardiomegaly & 0.90788 & 0.90909 & \textbf{0.91954} \\
        \hline
        Nodule & 0.79828 & 0.79695 & \textbf{0.80611} \\
        \hline
        Mass & 0.86191 & 0.86310 & \textbf{0.87315} \\
        \hline
        Hernia & 0.88305 & 0.84093 & \textbf{0.91723} \\
        \hline
        Average & 0.84239 & 0.83814 & \textbf{0.85433} \\
        \hline
    \end{tabular}
    \label{tab:ensemble_classical_dnn}
\end{table}

\vspace{-15pt} 
\begin{figure}[H]
    \centering
    \includegraphics[width=1\linewidth]{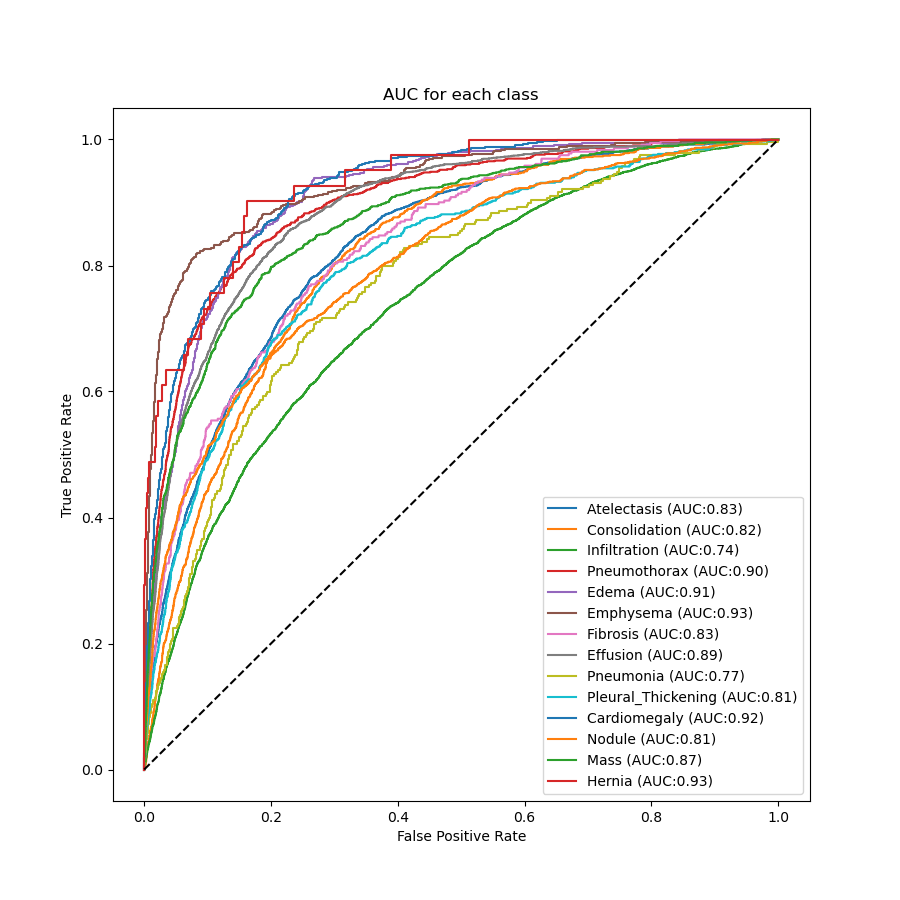}
    \caption{ROC curve for all 14 diseases displaying the True Positive against
False Positive rate, which illustrates the model’s capacity to differentiate effectively.}
    \label{fig:roc-curve}
\end{figure}





\begin{table*}[h]
    \centering
    \caption{COMPARISON OF AUROC WITH PREVIOUS WORK ON CHESTX-RAY14 DATASET}
    \begin{tabularx}{\textwidth}{|X|X|X|X|X|X|X|}
        \hline
        Disease & Ensemble (Ours) & CoAtNet (Ours) & $A^3$ Net\cite{wang_triple_2021} & ImageGCN\cite{mao_imagegcn_2022} & Wang et al.\cite{wang_chestx-ray8_2017} & Li et al.\cite{li_thoracic_2017} \\
        \hline
        Atelectasis & \textbf{0.83390} & 0.82313 & 0.779 & 0.802 & 0.716 & 0.800 \\
        Consolidation & \textbf{0.81575} & 0.80980 & 0.759 & 0.796 & 0.708 & 0.800 \\
        Infiltration & \textbf{0.74102} & 0.73105 & 0.710 & 0.702 & 0.609 & 0.700 \\
        Pneumothorax & \textbf{0.90164} & 0.89660 & 0.878 & 0.900 & 0.806 & 0.870 \\
        Edema & \textbf{0.91034} & 0.90185 & 0.855 & 0.883 & 0.835 & 0.880 \\
        Emphysema & 0.92946 & 0.92067 & \textbf{0.933} & 0.915 & 0.815 & 0.910 \\
        Fibrosis & 0.83347 & 0.81574 & \textbf{0.838} & 0.825 & 0.769 & 0.780 \\
        Effusion & \textbf{0.88977} & 0.88203 & 0.836 & 0.874 & 0.784 & 0.870 \\
        Pneumonia & \textbf{0.77648} & 0.76093 & 0.737 & 0.715 & 0.633 & 0.670 \\
        Pleural\_Thickening & \textbf{0.81270} & 0.80053 & 0.791 & 0.791 & 0.708 & 0.760 \\
        Cardiomegaly & \textbf{0.91954} & 0.90788 & 0.895 & 0.894 & 0.807 & 0.870 \\
        Nodule & \textbf{0.80611} & 0.79828 & 0.777 & 0.768 & 0.671 & 0.750 \\
        Mass & \textbf{0.87315} & 0.86191 & 0.834 & 0.843 & 0.706 & 0.830 \\
        Hernia & 0.91723 & 0.88305 & 0.938 & \textbf{0.943} & 0.767 & 0.770 \\
        Average & \textbf{0.85433} & 0.84239 & 0.826 & 0.832 & 0.738 & 0.804 \\
        \hline
    \end{tabularx}
    \label{tab:comparison_with_others}
\end{table*}

\subsection{Comparison with existing approaches}
A comparative analysis was conducted between our proposed model and previous competing models, as shown in Table \ref{tab:comparison_with_others}. The results demonstrated that our novel ensemble model produced superior outcomes. Notably, two exceptions occurred:  ImageGCN performed slightly better in the context of Hernia while $A^3$ Net\cite{wang_triple_2021} took the lead in relation to Emphysema and Fibrosis detection. As depicted by the ROC curve in Figure \ref{fig:roc-curve}, the ensemble model does well for all 14 diseases, excelling particularly in the case of Emphysema, yet showing comparatively less effectiveness for Infiltration. 

\section {Conclusion}
In this paper, we proposed an ensemble model for multi-label classification of chest X-rays (CXRs) using deep learning techniques. Firstly, we trained and evaluated several pure transformers, CNN, and hybrid models on the ChestX-ray14 dataset and found that the hybrid model CoAtNet performed the best individually, achieving an AUROC of 84.2\%. We also explored the performance of classical models like XGBoost and Random Forest when trained with feature vectors from our best-performing individual DNNs. Finally, we implemented a weighted average ensemble on the predictions of our DNNs using differential evolution to determine the optimal weight for each model. Our experiments show that our proposed ensemble model achieves better results than other state-of-the-art methods in this field, with a mean AUROC of 85.4\%. This demonstrates the potential of ensemble deep learning for improving the accuracy of automatic diagnosis of thoracic diseases from CXRs.

\bibliographystyle{ieeetr}
\bibliography{reference.bib}

\vspace{12pt}
\color{red}

\end{document}